\pdfoutput=1
\documentclass[11pt]{article}
\usepackage[final]{acl}

\usepackage{times}
\usepackage{latexsym}

\usepackage[T1]{fontenc}

\usepackage{kotex}
\usepackage{microtype}
\usepackage{inconsolata}
\usepackage{tcolorbox}

\usepackage{graphicx}
\usepackage{times,latexsym}
\usepackage{color}
\usepackage{colortbl}
\usepackage{hhline}
\usepackage[normalem]{ulem}

\usepackage{caption}
\usepackage{subcaption}
\usepackage{tabularx}
\usepackage{tikz}
\usepackage{pifont}
\usepackage{threeparttable}

\usepackage{xspace}
\usepackage{booktabs}  
\usepackage{boldline}
\usepackage{multirow}
\usepackage{subcaption}  
\usepackage{listings}
\usepackage{wrapfig}
\usepackage{enumitem}
\usepackage[section]{placeins}
\usepackage{soul}
\usepackage{threeparttable}
\usepackage{amsmath}  
\usepackage{amssymb}

\usepackage{hyperref} 
\usepackage{cleveref}

\newcommand\blfootnote[1]{%
  \begingroup
  \renewcommand\thefootnote{}\footnote{#1}%
  \addtocounter{footnote}{-1}%
  \endgroup
}

\definecolor{meta}{RGB}{1,0,245}    
\definecolor{init}{RGB}{234,51,247}   
\definecolor{aug}{RGB}{84,178,173}  

\title{Navigating the Path of Writing:\\Outline-guided Text Generation with Large Language Models}

\author{
 \textbf{Yukyung Lee\textsuperscript{1,\textdagger}}\quad
 \textbf{Soonwon Ka\textsuperscript{2}}\quad
 \textbf{Bokyung Son\textsuperscript{2}}\quad
 \textbf{Pilsung Kang\textsuperscript{3}}\quad
 \textbf{Jaewook Kang\textsuperscript{2}}
\\
 \textsuperscript{1}Boston University\quad
 \textsuperscript{2}NAVER AI Platform\quad
 \textsuperscript{3}Seoul National University
\\
\href{mailto:ylee5@bu.edu}{\texttt{ylee5@bu.edu}} \quad
\href{mailto:pilsung_kang@snu.ac.kr}{\texttt{pilsung\_kang@snu.ac.kr}} \\
\href{mailto:soonwon.ka@navercorp.com,bo.son@navercorp.com,jaewook.kang@navercorp.com}
{\texttt{\{soonwon.ka, bo.son, jaewook.kang\}@navercorp.com}}
}

\begin{document}

\maketitle
\blfootnote{\textsuperscript{\textdagger}Work done as a research intern at NAVER}

\begin{abstract}
Large Language Models (LLMs) have impacted the writing process, enhancing productivity by collaborating with humans in content creation platforms. However, generating high-quality, user-aligned text to satisfy real-world content creation needs remains challenging. We propose WritingPath, a framework that uses explicit outlines to guide LLMs in generating goal-oriented, high-quality text. Our approach draws inspiration from structured writing planning and reasoning paths, focusing on reflecting user intentions throughout the writing process. To validate our approach in real-world scenarios, we construct a diverse dataset from unstructured blog posts to benchmark writing performance and introduce a comprehensive evaluation framework assessing the quality of outlines and generated texts. Our evaluations with various LLMs demonstrate that the WritingPath approach significantly enhances text quality according to evaluations by both LLMs and professional writers.
\end{abstract}

\section{Introduction}

Writing is a fundamental means of structuring thoughts and conveying knowledge and personal opinions \cite{collins1980framework}. This process requires systematic planning and detailed review. \citet{hayes1980identifying} describes writing as a complex problem-solving process and explores how planning and execution interact in writing. That is, writing involves more than merely generating text; it encompasses developing a proper understanding of the topic, gathering relevant subject matter, and implementing thorough structuring.

\begin{figure}
    \centering
    \includegraphics[width=\columnwidth]{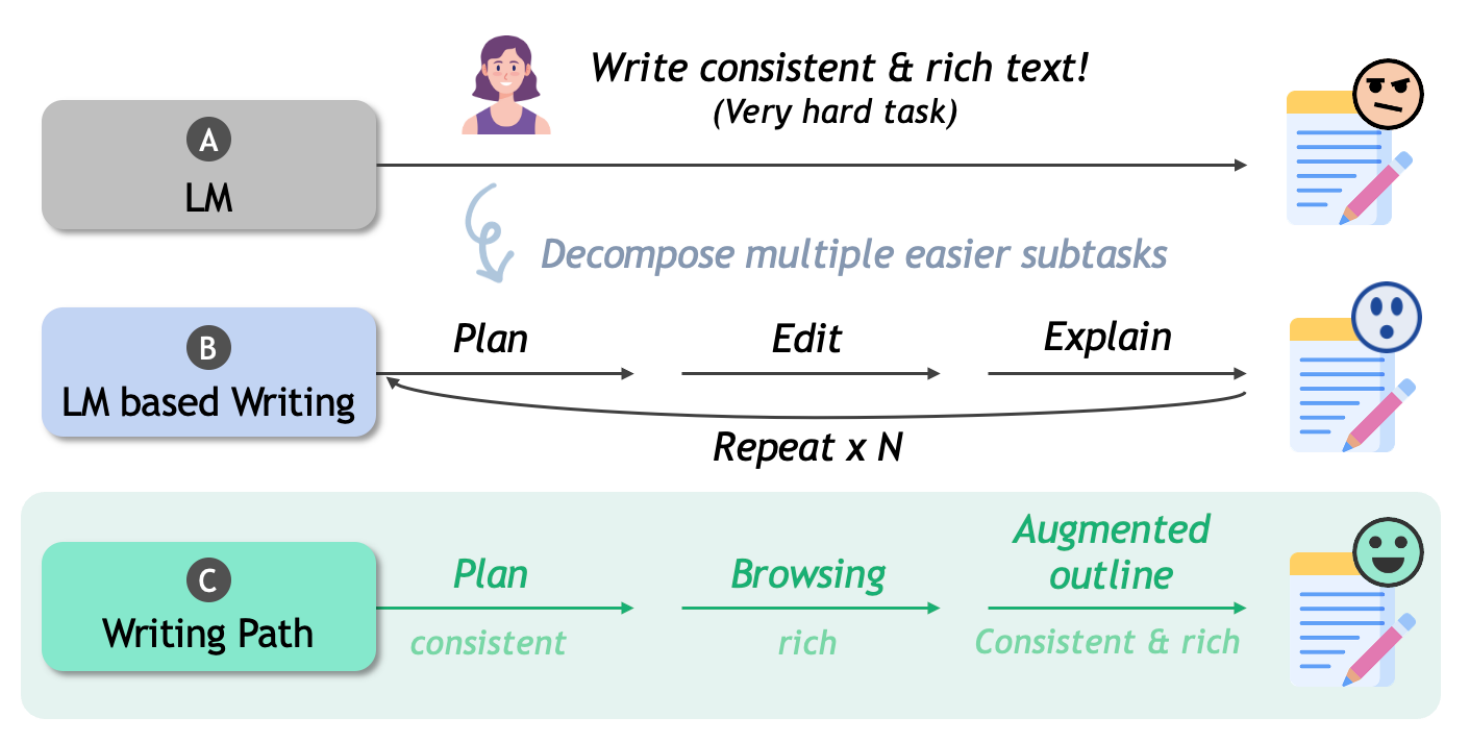}
    \caption{Comparative overview of writing approaches: (A) direct generation, (B) iterative writing involving planning, editing, and explaining, and (C) WritingPath method, which starts with a consistency-focused plan, incorporates information-rich browsing, and results in an augmented, consistent, and rich outline.}
    \label{fig:overview-of-wp}
\end{figure}
Recent advancements in Large Language Models (LLMs) have advanced the writing workflow, enhancing both its efficiency and productivity. One significant area of exploration is the collaborative use of LLMs in writing processes \cite{10.1145/3491102.3502030, mysore2023pearl}, as demonstrated by tools like Notion AI, Jasper, and Cohesive. The typical approach to incorporating LLMs involves the establishment of a writing plan and iterative improvement of interim outputs through revision \citet{schick2023peer, yang-etal-2022-re3}, as illustrated in Figure~\ref{fig:overview-of-wp} (b) with a focus on utilizing the generative capabilities of LLMs to improve fluency, consistency, and grammatical accuracy. While these tools support users in creating content more efficiently, there remains room for improvement in maintaining consistent quality that accurately aligns with specific user intentions in production environments \cite{wang2024weaver}.

To address this, we propose \textbf{WritingPath}, a methodology designed to incorporate user intentions such as desired topic, textual flow, keyword inclusion, and search result integration into the writing process. WritingPath emphasizes the importance of systematic planning and a clear outline from the early stages of writing. Inspired by the structured writing plan of \citet{hayes1980identifying} and the reasoning path of \citet{wei2022chain}, the WritingPath collects ideas and creates outlines that encapsulate the user's intentions before generating the final text. Furthermore, the initial outlines are further augmented with additional information through information browsing. Such a structured approach offers enhanced control over the text generation process and improves the quality of the content produced by LLMs.

We also utilize a multi-aspect writing evaluation framework to assess the intermediate and final productions from the WritingPath, offering a way to evaluate the quality of free-form text\footnote{Free-form text generation focuses on creating diverse texts tailored to specific information and user intentions, unlike story generation, which develops narratives with plots and characters} without relying on reference texts. Taking into account that conventional Likert scales (1-5 ratings) \cite{clark78c, hinkin1998brief} make it challenging to systematically compare and evaluate diverse writing outputs, particularly in creative tasks \cite{chakrabarty2023art}, our evaluation framework aims to provide more precise and reliable assessments for the outlines and final texts. For evaluation purposes, we construct a free-form blog text dataset incorporating a wide range of writing styles and topics from real users, including Beauty, Travel, Gardening, Cooking, and IT. Using this dataset, we evaluate how well the LLM outputs reflect the user's intentions. Applying the WritingPath to various LLMs shows significant performance gains across all evaluated models. These results validate that our approach enables the models to maintain a stronger focus on the given topic and purpose, ultimately generating higher-quality text that more accurately reflects user intentions. Furthermore, to validate real-world applicability, we applied WritingPath to a commercial writing platform for beta testing from October 2023 to March 2024. The deployment demonstrated its effectiveness in supporting real users with structured content creation across diverse writing needs.

The main contributions of this study can be summarized as follows:
\begin{itemize}
\item We propose WritingPath, a novel framework that enhances the ability of LLMs to generate high-quality and goal-oriented pieces of writing by using explicit outlines.
\item We customize a comprehensive evaluation framework that measures the quality of both the intermediate outlines and the final texts.
\item We construct a diverse writing dataset from unstructured blog posts across multiple domains, providing useful information such as aligned human evaluation scores, such as metadata that can be used as input for LLM-based writing tasks, and aligned human evaluation scores for the generated texts.
\item Our evaluation results indicate that the WritingPath markedly improves the quality of LLM-generated texts compared to methods that do not use intermediate outlines.
\end{itemize}

\section{Design of WritingPath}
\label{sec:Design of WritingPath}
We propose WritingPath, a systematic writing process to produce consistent, rich, and well-organized text with LLMs. Inspired by human writing processes, it consists of five key steps: metadata preparation, initial outline generation, information browsing, augmented outline creation, and final text writing (Figure~\ref{fig:2}). Each step is guided by a specific prompt configuration that aligns LLM output with specific step requirements. 

The core components of WritingPath are those that generate outlines as they establish a structured writing plan. Research suggests that a well-structured outline significantly impacts the quality of the written text \cite{sun-etal-2022-summarize, yang-etal-2022-re3, yang-etal-2023-doc}. The initial sketch is transformed into a detailed outline, including the flow, style, keywords, and relevant information from search results. This outline provides a clearer view of the final text to the LLMs. The specific steps are described as follows:

\begin{figure*}
    \centering
    \includegraphics[width=\linewidth]{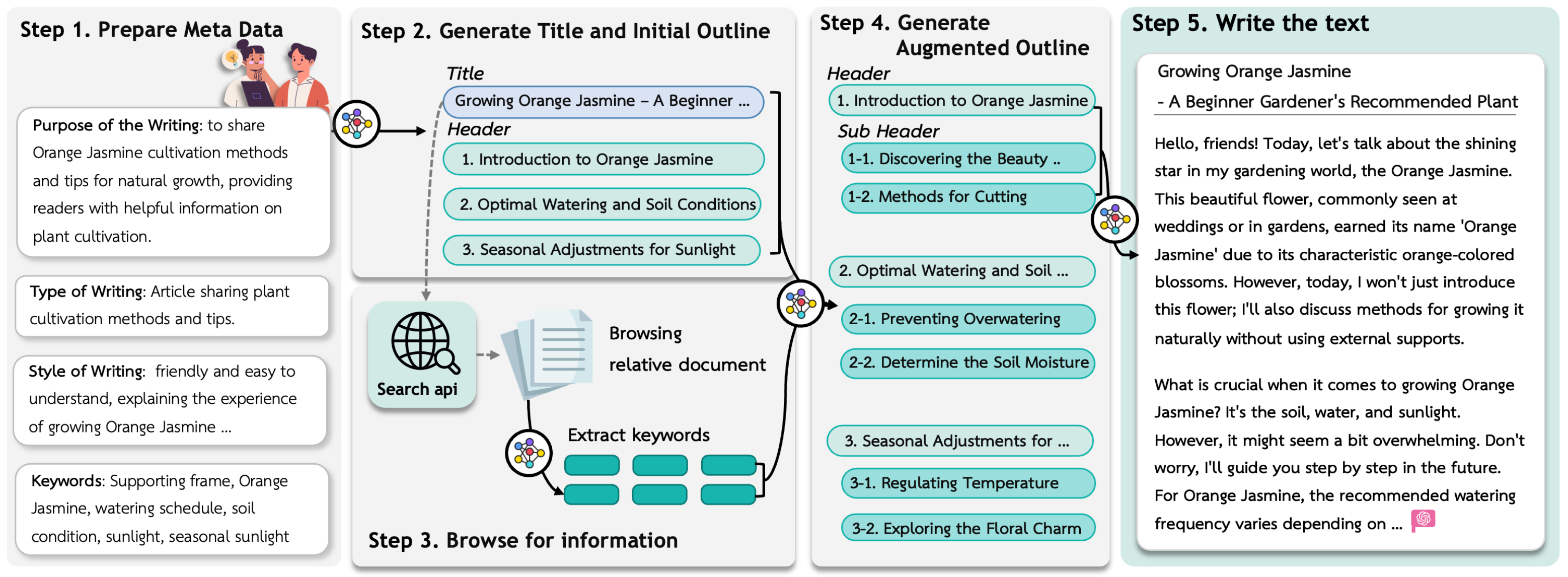}
    \caption{Main architecture of WritingPath, our proposed framework for guiding LLMs to generate high-quality text following a structured writing process. The WritingPath condenses text generation into five key steps. Inspired by human writing planning, it ensures alignment with specified writing goals.}
    \label{fig:2}
\end{figure*}

\textbf{Step 1: Prepare Meta Data} The first step establishes the writing direction and target reader using metadata $m$, which includes \lowercase\expandafter{\romannumeral1}) purpose, \lowercase\expandafter{\romannumeral2}) type, \lowercase\expandafter{\romannumeral3}) style, and \lowercase\expandafter{\romannumeral4}) keywords. To simulate this process, we converted human-written texts into metadata (see Section~\ref{sec:dataset} for details of the dataset).

\textbf{Step 2: Generate Title and Initial Outline} The second step generates the title $t$ and initial outline $O_{\text{init}}$ based on the metadata $m$ from step 1, using the LLM function $f_{\text{llm}}$ with a prompt configuration function $\phi_s$. Here, $s$ indicates the step index, and for step 2, the prompt configuration is $\phi_2$:
\vspace{-6pt}
\begin{equation}
t, O_{\text{init}} = f_{\text{llm}}(\phi_2(m)),
\vspace{-6pt}
\end{equation}
The initial outline $O_{\text{init}}$ consists of main headers $h_{i,0}$, where $i$ denotes the header sequence. This outline serves as the scaffolding of the text, organizing the main ideas and laying out the key points.

\textbf{Step 3: Browse for Information} The third step enriches the text by collecting additional information and keywords to reinforce the initial outline. We use the search function $f_{\text{search}}$ with the generated title $t$ as the query to retrieve the top-1 blog document, $D_{\text{sim}}$:
\vspace{-6pt}
\begin{equation}
D_{\text{sim}} = f_{\text{search}}(t)
\vspace{-6pt}
\end{equation}
In our implementation, we employ the NAVER search API\footnote{\href{https://developers.naver.com/docs/serviceapi/search/blog/blog.md}{https://developers.naver.com/docs/serviceapi}} to retrieve the top-1 document among similar blog posts. From the blog document, we extract keywords $K$ using the $f_{\text{llm}}$ with a prompt configuration $\phi_3$:
\vspace{-6pt}
\begin{equation}
K = f_{\text{llm}}(\phi_3(D_{\text{sim}})),
\vspace{-6pt}
\end{equation}
The extracted keywords constitute the additional information from the search results, leading to a more specific writing plan, improving the quality of the generated text.

\textbf{Step 4: Generate Augmented Outline} The fourth step refines the initial outline by adding subheadings and specific details to each section based on incorporating the keywords collected from the previous step. The augmented outline $O_{\text{aug}}$ is generated using the LLM function $f_{\text{llm}}$ with a prompt configuration $\phi_4$ that takes the title $t$, keywords $K$, and initial outline $O_{\text{init}}$ as inputs:
\vspace{-6pt}
\begin{align}
O_{\text{aug}} &= f_{\text{llm}}(\phi_4(t, k, O_{\text{init}})) \nonumber \\
&= \{ (h_{1,0}, \{h_{1,1}, h_{1,2}, \ldots\}), \nonumber \\
&\quad (h_{2,0}, \{h_{2,1}, h_{2,2}, \ldots\}), \nonumber \\
&\quad \ldots \}
\vspace{-6pt}
\end{align}
The resulting augmented outline, $O_{\text{aug}}$, comprises headers ($h_{i,0}$) and their corresponding subheaders ($h_{i,j}$), where $i$ denotes the header index, and $j$ indexes the subheaders. This detailed structure serves as a comprehensive writing plan, breaking down the text into manageable parts and providing clear direction for the content.

\textbf{Step 5: Write the Text} Finally, the text for each section $d^i$ is generated using the LLM function with a prompt configuration $\phi_5$ that takes the title $t$ and the corresponding section of the augmented outline $O_{\text{aug}}^i$ as inputs:
\vspace{-6pt}
\begin{equation}
d^i = f_{\text{llm}}(\phi_5(t, O_{\text{aug}}^i))
\vspace{-6pt}
\end{equation}
The final blog document $D$ is then compiled by concatenating all sections:
\vspace{-6pt}
\begin{equation}
D = \{d^1, d^2, \ldots, d^n\}
\vspace{-6pt}
\end{equation}
WritingPath organically connects all steps of the writing process, employing an outline to aggregate and manage diverse information, and assists users in producing high-quality writing. The prompts utilized for the WritingPath are detailed in Figure~\ref{fig:writing-path-prompt-1} and \ref{fig:writing-path-prompt-2}.
\section{Evaluation of WritingPath}
\label{sec:Evaluation of WritingPath}

Evaluating the effectiveness of WritingPath compared to existing writing support systems is challenging. Most previous studies do not directly utilize outlines in the writing process, resulting in a lack of systematic methods to assess outline quality. Even when outlines are used, evaluation relies only on human evaluation \cite{yang-etal-2023-doc, zhou2023recurrentgpt}. Moreover, current approaches heavily rely on human evaluation, which poses challenges for assessing full texts \cite{schick2023peer,yang-etal-2022-re3, lee2023evaluating}, as it requires evaluating multiple aspects of the written work. This challenge can arise in content creation workflows where scalable and consistent quality assessment helps maintain content standards.

To address these limitations, we propose an evaluation framework that combines human and automatic evaluation to assess the quality of generated outlines and final texts from multiple perspectives. This hybrid approach is designed to support real-world content creation workflows by combining systematic automated metrics with human assessment of nuanced writing aspects that require subjective judgment. The proposed method establishes clear evaluation criteria, enabling objective and reproducible validation of WritingPath's effectiveness as a writing support system.

\subsection{Outline Evaluation}
\subsubsection{Automatic Evaluation}
\label{sec:outline-auto-eval}

We adapt various metrics to evaluate the logical alignment, coherence, diversity, and repetition in outlines, following criteria established in linguistic literature \cite{van1977text, pitler-nenkova-2008-revisiting, tang-etal-2019-topic, elazar-etal-2021-measuring}. Logical alignment, assessed through NLI-based methods, ensures that headers and subheaders are logically connected. Coherence evaluates thematic uniformity across sections, while diversity measures the breadth of topics covered. Repetition is analyzed to minimize redundancy and improve information efficiency. Note that coherence and diversity exhibit a trade-off relationship; maintaining coherence while covering a wide range of topics is essential to ensure the effectiveness of the outline in guiding the writing process. Detailed evaluation definitions are available in Appendix~\ref{app:outline-auto-eval}.

\subsubsection{Human Evaluation}
\label{sec:outline-human-eval}

In addition to automatic evaluation metrics, we conduct a human evaluation to assess aspects of the generated outlines that are difficult to capture solely with automatic measures. These aspects include cohesion, natural flow, and redundancy. For augmented outlines, we also evaluate the usefulness of added information and overall improvement compared to the initial outline. Detailed evaluation definitions are available in Appendix~\ref{app:outline-human-eval}.

\begin{figure}[t]
    \centering
    \includegraphics[width=\columnwidth]{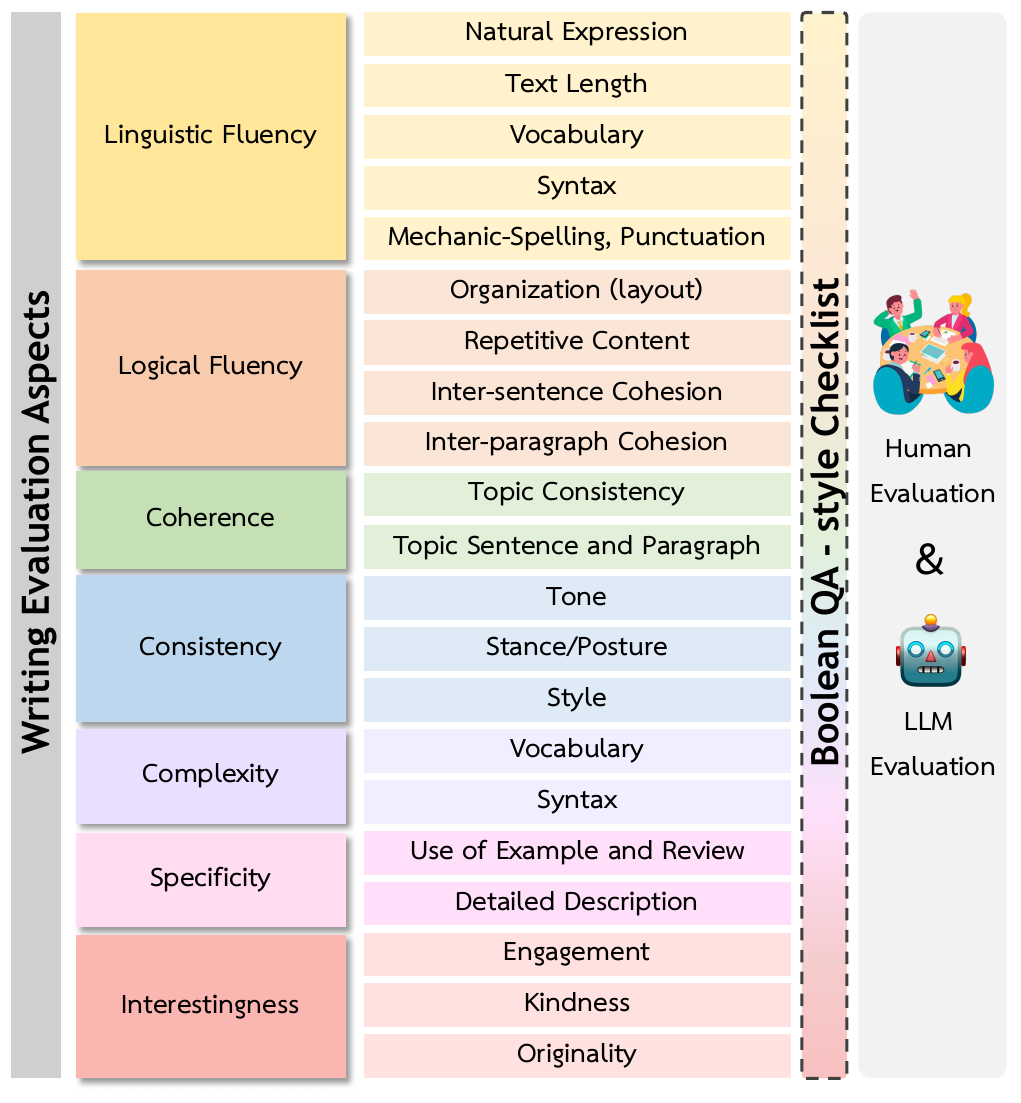}
    \caption{Breakdown of the seven key aspects used in writing evaluation, each with corresponding sub-aspects, employed in a Boolean QA-style checklist for human and LLM evaluation. This comprehensive framework ensures a multi-dimensional analysis of text quality.}
    \label{fig:writing-aspects}
\end{figure}

\begin{table*}[ht]
    \centering
    \resizebox{\textwidth}{!}{%
    \footnotesize
        \begin{tabular}{l|cccc|cccc} 
        \toprule
        \textbf{Model}
        & \multicolumn{4}{c|}{\textbf{Automatic Evaluation}} & \multicolumn{4}{c}{\textbf{Human Evaluation}} \\
        \midrule
        \textbf{Aspects} & \textbf{Logical Alignment} & \textbf{Coherence} & \textbf{Diversity} & \textbf{Repetition} & \textbf{Cohesion} & \textbf{Natural Flow} & \textbf{Diversity} & \textbf{Redundancy} \\
        Metrics& NLI ($\uparrow$) & UCI ($\uparrow$) / NPMI ($\uparrow$) & Topic Diversity ($\uparrow$) & Self-BLEU ($\downarrow$) & ($\uparrow$) & ($\uparrow$) & ($\uparrow$) & ($\uparrow$) \\
        \midrule
        Eval Level & Header-Subheader & Outline & Outline & Outline & Outline & Outline & Outline & Outline \\
        \midrule
        \multicolumn{1}{l|}{GPT-3.5} & \multicolumn{4}{c|}{\multirow{2}{*}{}} & \multicolumn{4}{c}{\multirow{2}{*}{}} \\
        \multicolumn{1}{r|}{\makebox[2cm][r]{\textit{initial}}} & - & 0.60 / 0.31 & 0.60 & 32.03 & \textbf{3.38} & 2.70 & 2.77 & 2.73 \\
        \multicolumn{1}{r|}{\makebox[2cm][r]{\textit{augmented}}} & 0.61 & \textbf{1.33} / \textbf{0.51} & \textbf{0.61} & \textbf{17.33} & 3.15 & \textbf{2.78} & \textbf{3.54} & \textbf{3.13} \\
        \midrule
        \multicolumn{1}{l|}{GPT-4} & \multicolumn{4}{c|}{\multirow{2}{*}{}} & \multicolumn{4}{c}{\multirow{2}{*}{}} \\
        \multicolumn{1}{r|}{\makebox[2cm][r]{\textit{initial}}} & - & 0.80 / 0.49 & 0.67 & 24.81 & \textbf{3.40} & 2.86 & 3.06 & 2.86 \\
        \multicolumn{1}{r|}{\makebox[2cm][r]{\textit{augmented}}} & 0.66 & \textbf{1.61} / \textbf{0.52} & \textbf{0.68} & \textbf{13.12} & \textbf{3.40} & \textbf{2.98} & \textbf{3.74} & \textbf{3.43} \\
        \midrule
        \multicolumn{1}{l|}{HyperCLOVA X} & \multicolumn{4}{c|}{\multirow{2}{*}{}} & \multicolumn{4}{c}{\multirow{2}{*}{}} \\
        \multicolumn{1}{r|}{\makebox[2cm][r]{\textit{initial}}} & - & 0.75 / 0.41 & 0.74 & 18.04 & \textbf{3.47} & 2.96 & 2.82 & 3.22 \\
        \multicolumn{1}{r|}{\makebox[2cm][r]{\textit{augmented}}} & 0.67 & \textbf{1.82} / \textbf{0.54} & \textbf{0.75} & \textbf{11.50} & 3.41 & \textbf{3.48} & \textbf{3.93} & \textbf{3.79} \\
        \bottomrule
        \end{tabular}}
    \caption{Automatic and human evaluations on the quality of initial and augmented outlines from GPT-3.5, GPT-4, and HyperCLOVA X. Bold indicates the best result within a model.}
    \label{tab:outline-eval-results}
\end{table*}

\subsection{Writing Evaluation}

Traditional evaluation metrics such as Likert scales are not well-suited for assessing creative tasks like long story generation \cite{chakrabarty2023art}. Acknowledging the need for more specific writing evaluation methods, we employ CheckEval \cite{lee2024checkeval} to assess writing quality\footnote{\citet{lee2024checkeval} reports a 0.65 spearman correlation between human and LLM evaluations for dialogue, which surpasses G-Eval \cite{liu2023gpteval}. This demonstrates CheckEval's potential as a reliable method for evaluating model-generated text quality.}. CheckEval decomposes the evaluation aspects into more granular sub-questions, forming a detailed checklist. These aspect-based checklists can make performance evaluations by either humans or LLMs more fine-grained. Moreover, by explicitly capturing the evaluator's reasoning behind each rating, this approach enhances the explainability of the evaluation process. To adapt CheckEval, we identified 7 aspects and selected relevant sub-aspects for each. We formulated them as binary (Yes/No) questions. This resulted in a checklist-style evaluation sheet for each sub-aspect, enabling an intuitive and structured assessment of the generated texts. The prompts utilized for the writing evaluation are detailed in Figure~\ref{fig:writing-evaluation-prompt}. The evaluation criteria were selected based on prior linguistics research \cite{WOLFE199783, KNOCH201181, Lee2019BestPF, Celikyilmaz2020EvaluationOT, Chhun2022OfHC,10.1145/3485766, VANDERLEE2021101151} and finalized through a review and refinement process involving 6 writing experts. Details of the evaluation criteria are in Figure~\ref{fig:writing-aspects}, and the instructions and checklist used during the evaluation process are presented in Table~\ref{tab:eval-principles}.

\section{Experimental Setting}
\subsection{Dataset}
\label{sec:dataset}

In this study, we constructed a Korean dataset based on real user-written blog posts to assess the effects of the WritingPath in real content creation scenarios. The dataset covers five domains frequently handled in content creation:: travel, beauty, gardening, IT, and cooking. We created a total of 1,500 posts for each model, resulting in 4,500 instances in total. For human evaluation, we randomly sampled 10\% of the outlines and texts and assessed their scores. Final texts were evaluated by human experts, aligning model outputs with professional quailty standards. Details of the dataset are in Appendix~\ref{app:details of dataset}.

\begin{figure}[t]
    \centering
    \includegraphics[width=\columnwidth]{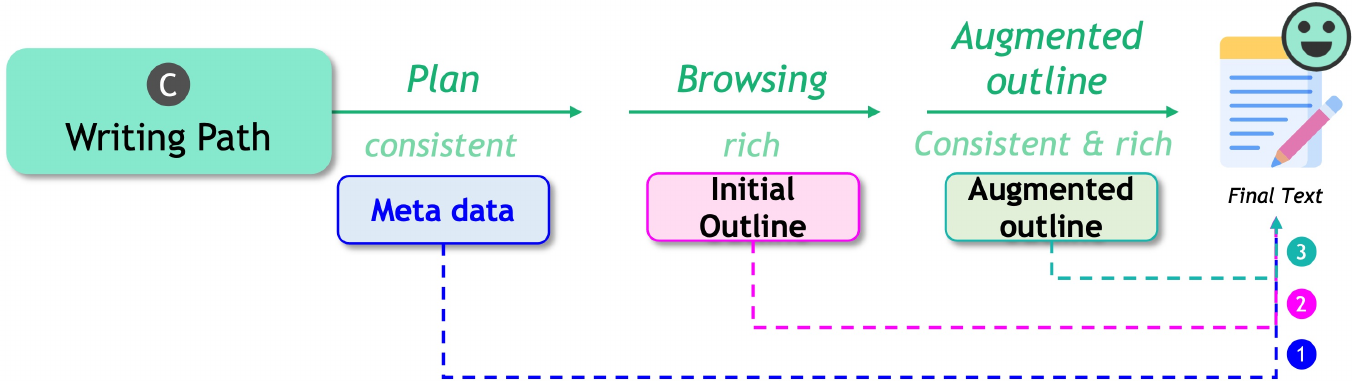}
    \caption{Overview of the main analysis steps in the WritingPath framework, covering meta-data only, initial outline, and augmented outline scenarios, respectively.}
    \label{fig:writing_path_analysis}
\end{figure}

\subsection{Model}
We conducted experiments using three models: GPT-3.5-turbo \cite{brown2020language}, GPT-4 \cite{achiam2023gpt}, and HyperCLOVA X \cite{yoo2024hyperclova}\footnote{\texttt{gpt-3.5-turbo}, \texttt{gpt-4-0125}, \texttt{HCX-003}}. For evaluation, we used GPT-4-turbo\footnote{\texttt{gpt-4-turbo}; we chose GPT-4-turbo as the evaluation model because of its best performance at the time of this study.}. Additionally, we attempted to adapt the WritingPath approach to open-source models, including Llama2, Orion, and KoAlpaca. However, their outputs did not meet the quality standards necessary for fair comparison, and they were excluded from our analysis.

\subsection{Human Evaluation}
We conducted two separate human evaluation processes, involving a total of 12 carefully selected evaluators. For outline evaluations, which are relatively simple and short, we employed 6 native Korean speakers with experience in LLM. For the more detailed and rigorous writing evaluation, we recruited 6 professional writers and teachers as writing experts, each with over 10 years of expertise in Korean writing.
\section{Experimental Results}
\subsection{Effectiveness of WritingPath}
\label{sub-sec:Effectiveness of WritingPath}

\begin{figure}[t]
    \captionsetup[subfigure]{labelformat=empty,aboveskip=-1pt,belowskip=-1pt}
    \centering
    \begin{subfigure}{0.493\columnwidth}
        \centering
        \includegraphics[width=\textwidth]{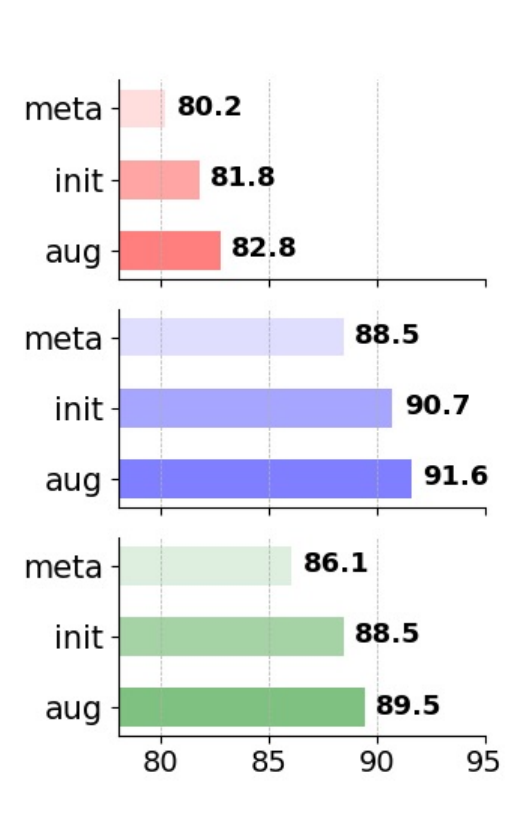}
        \caption{(a) LLM Evaluation}
        \label{fig:average-writing-llm-eval-analysis}
    \end{subfigure}
    \begin{subfigure}{0.493\columnwidth}
        \centering
        \includegraphics[width=\textwidth]{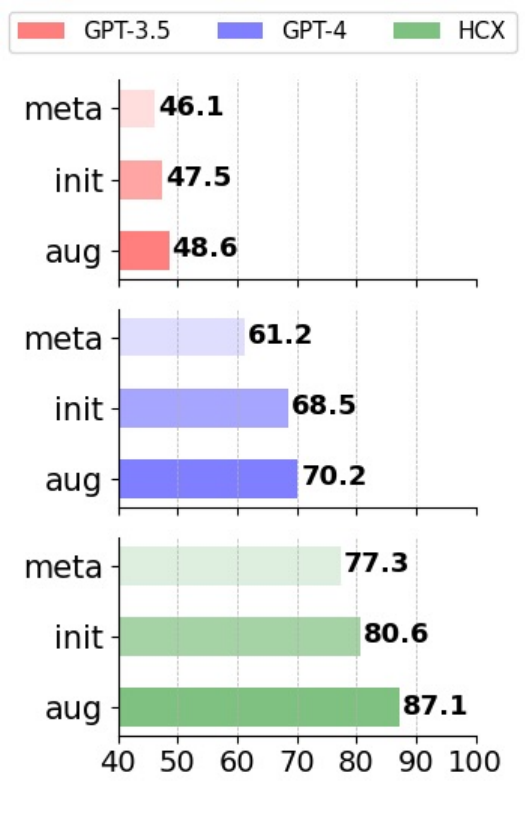}
        \caption{(b) Human Evaluation}
        \label{fig:average-writing-human-eval-analysis}
    \end{subfigure}
    \caption{Main analysis steps on writing evaluation results by (a) LLM and (b) Human Evaluation.}
    \label{fig:average-writing-eval-analysis}
\end{figure}

To verify that going through the WritingPath improves the final writing quality, we designed an analysis incorporating three cases (Figure~\ref{fig:writing_path_analysis}): {\color{meta}\ding{202}} writing from metadata, {\color{init}\ding{203}} writing from the initial outline, {\color{aug}\ding{204}} writing from the augmented outline, where this final case corresponds to the complete WritingPath pipeline.

Figure~\ref{fig:average-writing-eval-analysis} shows results from both (a) LLM and (b) human evaluation using CheckEval. Both consistently show progressive improvement as more components of the WritingPath are incorporated, while the model rankings are in different order between the two evaluation methods\footnote{In the LLM evaluation, GPT-4 outperforms HyperCLOVA X, whereas the opposite trend is observed in human evaluations. These differences may be due to the use of GPT-4-turbo as the evaluation model and the self-enhancement bias discussed in \citet{Zheng2023JudgingLW}.}. Specifically, The results show that using the augmented outline (aug) leads to better writing quality compared to using only metadata (meta), indicating that the quality of writing improves significantly when the full WritingPath pipeline is employed. Furthermore, the augmented outline (aug) outperforms the initial outline (init), indicating that the content enrichment process further enhances writing quality. For a comprehensive analysis of writing quality, including human evaluation results for the final text across models, detailed improvement of text quality through the WritingPath, and Kendall tau correlations between various writing aspects and overall text quality, see Appendix~\ref{app:further-analysis}.

\subsection{Outline Evaluation} 
Section~\ref{sub-sec:Effectiveness of WritingPath} showed that using the augmented outline in the WritingPath pipeline led to better performance compared to using only the initial outline or metadata. To assess not only the impact of initial and augmented outlines on the quality of the final writing but also any differences in quality at the outline stage itself, we evaluated the initial and augmented outlines independently. 

\begin{figure}[t]
    \centering
    \includegraphics[width=\columnwidth]{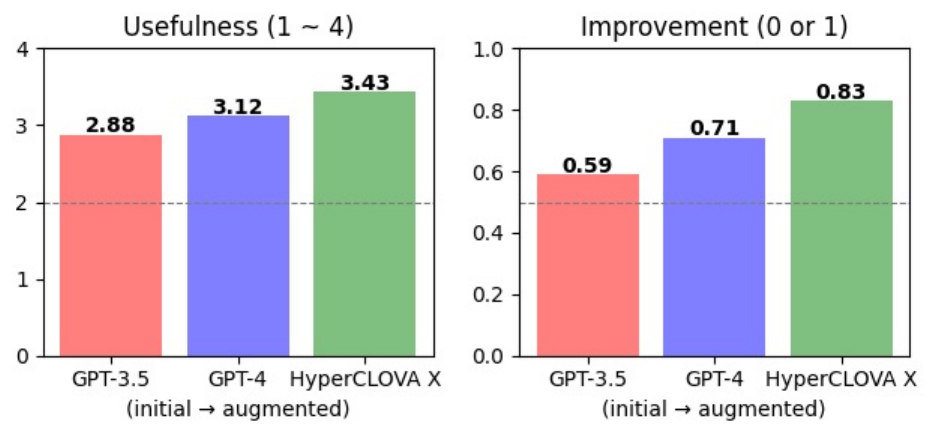}
    \caption{Evaluation of augmented outlines showing all models surpass the effectiveness threshold with scores in Usefulness above 2 and Improvement over 0.5, indicating universal enhancements from the initial outlines.}
    \label{fig:augmented-outline-improvement}
\end{figure}

\textbf{Automatic Evaluation}
To see the effects of the outline augmentation module, we conducted automatic evaluations on the initial and augmented outlines using criteria described in Section~\ref{sec:outline-auto-eval}. The results in Table~\ref{tab:outline-eval-results} show significant improvements in Coherence and Repetition aspects for the augmented outlines compared to the initial ones, indicating that the outline augmentation process enhances content consistency and reduces unnecessary repetition. Notably, although Diversity and Coherence are often considered trade-offs, the augmented outlines in our study maintained Diversity while improving Coherence. This suggests that the outline expansion module can increase consistency without compromising content diversity. Detailed performance across various domains is in Table~\ref{tab:detailed-outline-llm-eval-results}.

\textbf{Human Evaluation}
As described in Section~\ref{sec:outline-human-eval}, we conducted human evaluations to assess the cohesion, natural flow, diversity, and redundancy of initial and augmented outlines. The augmented outlines demonstrated significant improvements in all aspects except cohesion, which slightly declined or remained stable. Nevertheless, the overall performance of the augmented outlines surpassed that of the initial outlines. Further evaluations of the augmented outlines were conducted on usefulness and improvement, which indicated the extent of useful information added and overall quality enhancement compared to the initial outlines. As shown in Figure~\ref{fig:augmented-outline-improvement}, all models demonstrated improvements in both metrics, validating the power of the browsing step. Detailed performance across various text domains is in Table~\ref{tab:detailed-outline-human-eval-results}.
\section{Real-World Deployment}
\begin{figure*}[t]
    \centering
    \includegraphics[width=0.95\textwidth]{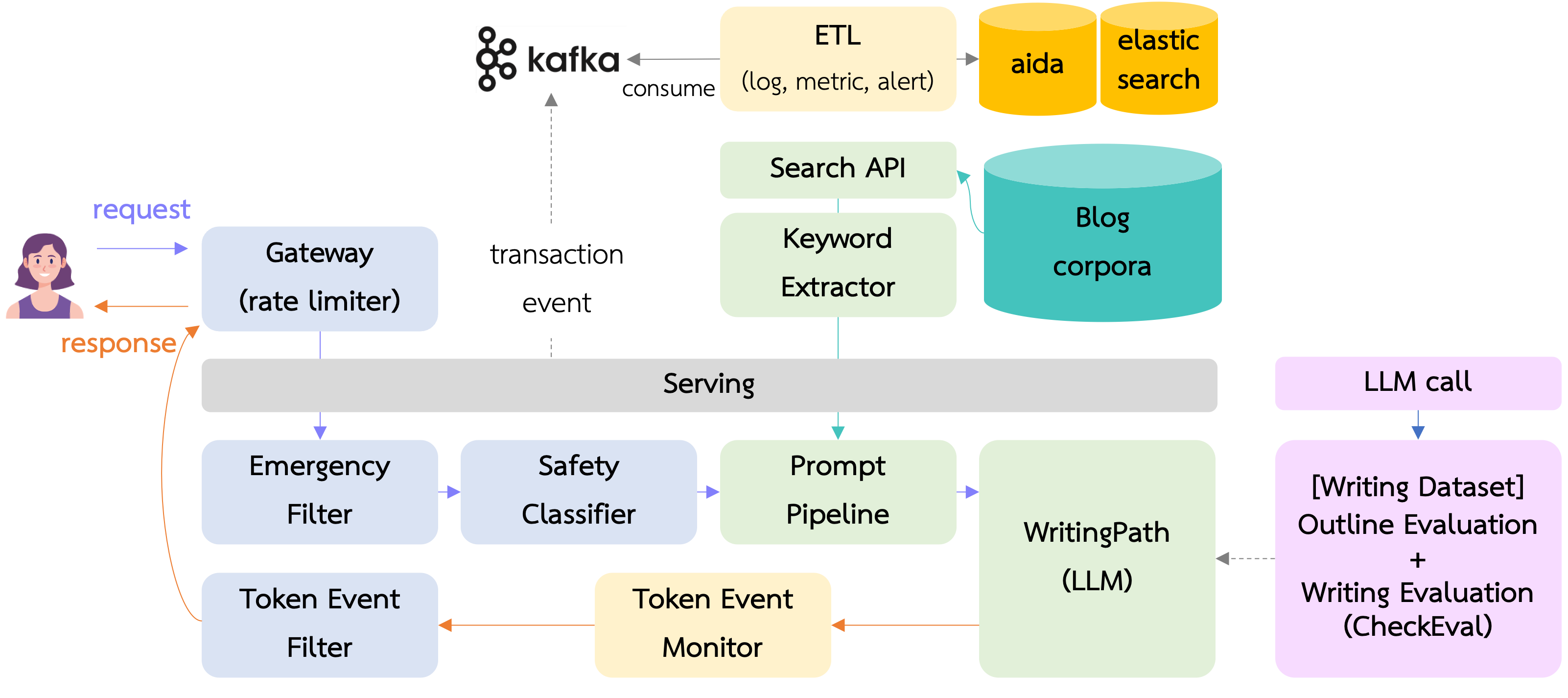}
    \caption{Real-world deployment pipeline of WritingPath.}
    \label{fig:real-world-pipeline}
\end{figure*}

WritingPath was integrated into a commercial blogging platform as a writing assistance feature and tested for six months. In the service environment, additional considerations such as safety filtering and content quality control measures were necessary for reliable content generation. The system architecture of CLOVA for Writing by NAVER is depicted in Figure \ref{fig:real-world-pipeline}.

The serving pipeline integrates multiple components for reliable service operation. It integrates user request handling, content filtering, Kafka pipeline, and retrieval. Requests pass through a Gateway with rate limiting and are filtered for harmlessness. Specifically, the system includes emergency filtering and safety classification before passing requests to WritingPath. Additionally, a token event monitoring system tracks model usage, followed by token event filtering over output anomalies.
\section{Conclusion}
We introduced WritingPath, a framework that enhances the ability of LLMs to generate high-quality and goal-oriented writing by employing explicit outlines. Designed for real-world content creation, our approach uses structured guidelines from the early stages to ensure consistent quality control.

We verified the impact of WritingPath by conducting a comprehensive evaluation that incorporates automatic and human evaluations covering a wide range of aspects. Our experimental results demonstrate that texts generated following the full WritingPath approach, which includes the use of augmented outlines, exhibit superior performance compared to texts produced using only initial outlines or without any intermediate outlines. We also proposed a framework for assessing the WritingPath's intermediate outlines, which found that augmented outlines have better inherent quality than initial outlines, demonstrating the importance of outline augmentation steps. We hope that this work will contribute to the research and development of more reliable AI-assisted writing solutions.
\section{Acknowledgments}

This research project was conducted as part of the NAVER HyperCLOVA-X and CLOVA for Writing projects. We express our gratitude to Nako Sung for his thoughtful advice on writer LLMs and to the NAVER AX-SmartEditor team. Additionally, we would like to thank the NAVER Cloud Conversational Experience team for their practical assistance and valuable advice in creating the blog dataset. 
We appreciate Jaehee Kim, Hyowon Cho, Keonwoo Kim, Joonwon Jang, Hyojin Lee, Joonghoon Kim, Sangmin Lee, and Jaewon Cheon for their invaluable feedback and evaluation. We also thank DSBA NLP Group members for their comments on the paper.

\bibliography{custom}

\clearpage
\appendix
\section{Related Work}
\label{related_work}

\subsection{Collaborative Writing with Language Models}
Recent works that explore collaboration with LLMs during the writing process can be categorized into two aspects: 1) Outline Planning and Draft Generation, and 2) Recursive Re-prompting and Revision.

\textbf{Outline Planning and Draft Generation} involves incorporating the writer's intents and contextual information into LLM prompts to create intermediate drafts. Dramatron \citep{dramatron} is a system for collaborative scriptwriting that automatically generates outlines with themes, characters, settings, flows, and dialogues. DOC \citep{yang-etal-2023-doc} improves the coherence of generating long stories by offering detailed control of their outlines, including analyses of generated outlines and suggestions for revisions to maintain consistent plot and style. 

Building on these works, our WritingPath mimics the human writing process by structuring it into controllable outlines. While our approach shares similarities with DOC in terms of utilizing outlines, we diverge from focusing solely on story generation and propose a novel outline generation process that incorporates external knowledge through browsing. Our aim is to sophisticatedly control machine-generated text across a wide range of writing tasks.

\textbf{Recursive Reprompting and Revision} technique extends the potential of LMs to assist not only with draft generation but also with editing and revision processes. This approach employs LLM prompt chains such as planning - drafting - reviewing - suggesting revisions in an iterative fashion to enhance the quality of written content. Re3 \citep{yang-etal-2022-re3} introduces a framework for maintaining the long-range coherence of draft generation. It operates separate rewriter and edit modules in its prompt chain to check and refine plot relevance and long-term factual consistency. PEER \citep{schick2023peer} proposes a recursive revision framework based on the concept of self-training, where the model autonomously selects the editing operations for revision and provides explanations for the modifications it makes. RECURRENTGPT \cite{zhou2023recurrentgpt} utilizes a recursive, language-based mechanism to simulate LSTM \cite{10.1162/neco.1997.9.8.1735}, enabling the generation of coherent and extended texts. While these works are relevant to collaborative writing with LMs, direct comparisons with our approach are unfeasible. These studies focus on specific tasks like story generation, requiring task-specific training and datasets, which are unavailable in Korean for our writing tasks.

Our WritingPath differs from previous works in its goals for utilizing LLMs in the writing process. Instead of relying on an ad-hoc recursive writing structure that may be inefficient, we establish a systematic writing plan that guides the generation process from the very beginning. Furthermore, we focus on free-form text generation rather than story generation and do not require separate training for writing, planning, or editing.

\subsection{Integrating External Information}
Existing approaches have explored various methods to inject external knowledge into LLMs to improve their performance on text-generation tasks. For instance, Retrieval-Augmented Generation (RAG) \cite{lewis2020retrieval}, and Toolformer \cite{schick2024toolformer} have developed techniques to connect LLMs with external search tools, enabling them to gather relevant information and generate more informative and accurate responses. However, despite these contributions to improving LLMs' access to information \cite{asai2024selfrag}, they inherently fall short of fully reflecting the diversity and complexity of the writing process \cite{chakrabarty2023art}. 

Our work distinguishes itself from previous approaches by focusing on emulating the modern writing planning process. With this structured approach, an LLM can efficiently produce high-quality text, significantly contributing to improving the control and quality of the generated text.

\subsection{Writing Evaluation}
It is well-known that supervised metrics such as ROUGE and BLEU are ill-suited for evaluating natural language generation output, especially for open-ended writing tasks. Traditionally, such evaluation has depended on rubric-based human evaluation, which is a costly and time-consuming task \cite{assessing-writing}. Recent advancements in LLMs have led to the exploration of new paradigms that utilize LMs for evaluating LM-generated text \cite{gralinski-etal-2019-geval, fu2023gptscore}. However, to effectively assess free-form text, a more customized and interactive evaluation framework is needed.

We utilize CheckEval \cite{lee2024checkeval}, a fine-grained and explainable evaluation framework, to assess free-form text writing. By customizing a checklist with specific sub-questions for each writing aspect, we provide a more reliable and accurate means of evaluating writing quality.

\section{Details of Dataset}
\label{app:details of dataset}
We selected 20 blog posts for each domain\footnote{\href{https://blog.naver.com/}{https://blog.naver.com/}}, resulting in 100 seed data points. For seed data construction, we generated metadata, including purpose, topic, keywords, and expected reader, based on the title and content of the blog posts. This metadata is the input to the WritingPath, helping the model understand the context of the post and generate relevant outlines and text. We created a test dataset of 1,100 instances per model under evaluation using the seed data. Each data point includes the outputs of each WritingPath step: an outline, additional information, an augmented outline, and the final text. With analysis experiments as well, we generated a total of 1,500 posts for each model, resulting in 4,500 instances in total. For human evaluation, we randomly sampled 10\% of the outlines and texts and assessed their scores. The final texts were evaluated by human experts, and the dataset aligns the generated outputs from three models with the human scores.

\section{Details of Evaluation}
\subsection{Compensation Details}
Outline evaluators were compensated with a 6,000 KRW ($\approx$ 4.2 USD) gift card for their 30-minute participation. And writing experts were compensated at a rate of 9,000 KRW ($\approx$ 6.6 USD) per one-writing sample.

\subsection{Automatic Evaluation - Outline}
\label{app:outline-auto-eval}
\begin{itemize}
    \item Logical alignment: Based on \citet{chen-eger-2023-menli}, we utilize Natural Language Inference (NLI) which examines whether the headers and subheaders within an outline logically connect, ensuring the structural integrity necessary for coherent argumentation\footnote{we utilize \texttt{gpt-4-turbo} for NLI evaluation}. 
    \item Coherence: Through Topic Coherency metrics such as NPMI \cite{stevens-etal-2012-exploring} and UCI \cite{lau-etal-2014-machine}, this aspect assesses the thematic uniformity across the sections of outline, verifying a consistent narrative.
    \item Diversity: We measure the breadth of topics addressed by applying Topic Diversity metrics \cite{dieng-etal-2020-topic}, aiming to ensure that the content of outline is comprehensive and varied.
    \item Repetition: Self-BLEU \cite{zhu2018texygen} is used to gauge the degree of redundancy within the outline, prioritizing efficiency in information presentation by minimizing repetition.
\end{itemize}

\begin{table*}[ht]
    \centering
    \resizebox{0.95\textwidth}{!}{%
   \footnotesize
    \begin{tabular}{l|ccccccc|c}
        \toprule
        \textbf{Model} & \textbf{Linguistic Fl.} & \textbf{Logical Fl.} & \textbf{Coh.} & \textbf{Cons.} & \textbf{Comple.} & \textbf{Spec.} & \textbf{Int.} & \textbf{Overall} \\
        & binary & binary & binary & binary & binary & binary & binary & binary \\
        \midrule
        GPT-3.5     & 51.66 & 31.14 & 46.29 & 88.11 & 66.43 & 21.14 & 35.14 & 48.56 \\ 
        \midrule
        GPT-4     & 68.00 & 60.57 & 72.86 & 89.26 & 80.29 & 54.14 & 66.29 & 70.20 \\
        \midrule
        HyperCLOVA X     & \textbf{89.71} & \textbf{84.46} & \textbf{91.14} & \textbf{98.06} & \textbf{92.57} & \textbf{74.00} & \textbf{80.00} & \textbf{87.13} \\
        \bottomrule
    \end{tabular}}
    \caption{Human evaluation results for writing quality of final text (aug) across models.}
    \label{tab:writing-human-eval-results}
\end{table*}

\begin{figure*}
    \centering
    \includegraphics[width=0.95\textwidth]{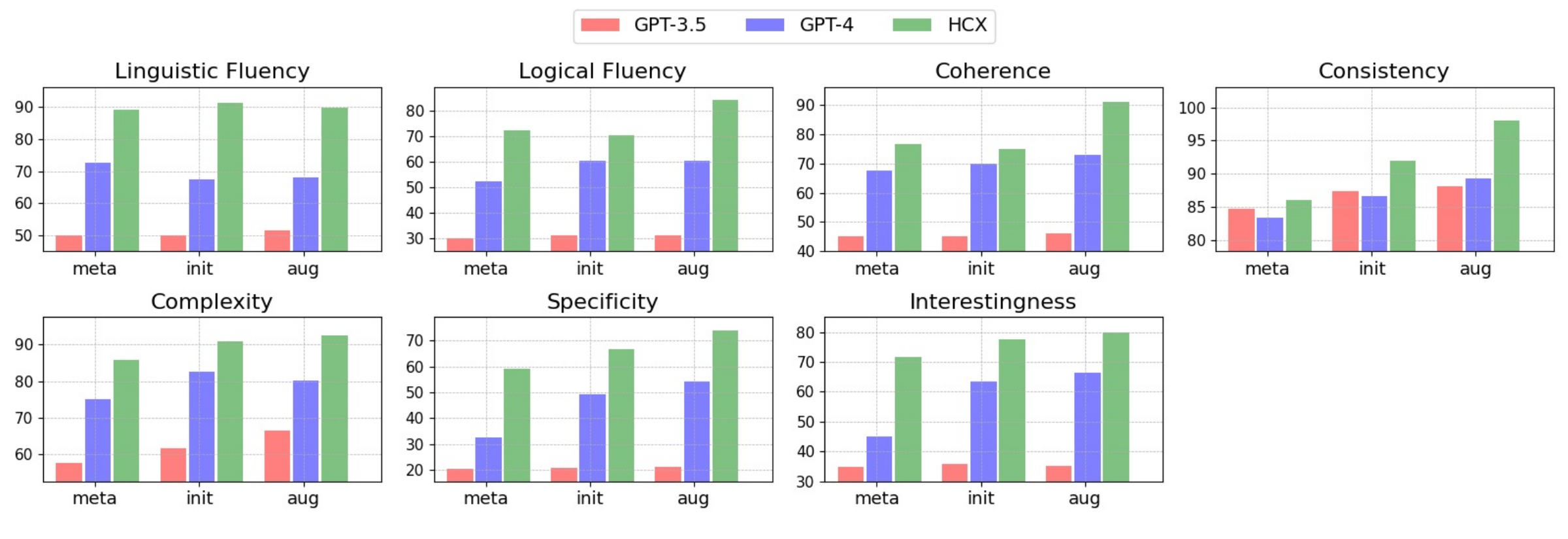}
    \caption{Human evaluation results for writing quality (meta, init, aug) over various CheckEval aspects.}
    \label{fig:writing-human-eval-analysis}
\end{figure*}

\subsection{Human Evaluation - Outline}
\label{app:outline-human-eval}
The human evaluation criteria are based on aspects considered in previous studies on text coherence, relevance, and quality assessment \cite{yang-etal-2022-re3, yang-etal-2023-doc,zhou2023recurrentgpt, ke-etal-2022-ctrleval}. For both initial and augmented outlines, the human evaluation is performed on the following five aspects, using a 1-4 point scale:
\begin{itemize}
    \item Cohesion: Evaluates whether the title and outline are semantically consistent.
    \item Natural Flow: Assesses whether the outline flows in a natural order.
    \item Diversity: Evaluates whether the outline consists of diverse topics.
    \item Redundancy: Assesses whether the outline avoids semantically redundant content.
\end{itemize}
Furthermore, we use two additional aspects for evaluating the augmented outline:

\begin{itemize}
    \item Usefulness of Information: Assesses whether the augmented outline provides useful information beyond the initial outline.
    \item Improvement: Evaluates whether significant improvements have been made in the augmented outline compared to the initial outline, using a binary scale.
\end{itemize}

\section{Further Analysis of Writing Quality}
\label{app:further-analysis}
To further analyze the quality of the text generated through the complete WritingPath pipeline, we conducted a human evaluation based on the CheckEval framework. The results are presented in Table~\ref{tab:writing-human-eval-results}. The analysis by six writing experts showed that GPT-4 and HyperCLOVA X generally performed better than GPT-3.5 in terms of writing quality. HyperCLOVA X exhibited higher scores in specificity compared to other models, which is consistent with the findings reported in KMMLU \cite{son2024kmmlu} regarding the advantages of language-specific models. Detailed performance metrics across various domains and further LLM evaluations can be found in Table~\ref{tab:detailed-writing-llm-eval-results},~\ref{tab:detailed-writing-human-eval-results}. Furthermore, We consider seven key aspects (Section~\ref{fig:writing-aspects}) for evaluating the quality of writing. CheckEval’s binary responses for each aspect allow for identifying the specific factors contributing to the assessments. We found that logical fluency, coherence, consistency, and specificity significantly contribute to the improvement of text quality through the WritingPath (Figure~\ref{fig:writing-human-eval-analysis}). 

\begin{figure}[t]
    \centering
    \includegraphics[width=0.95\columnwidth]{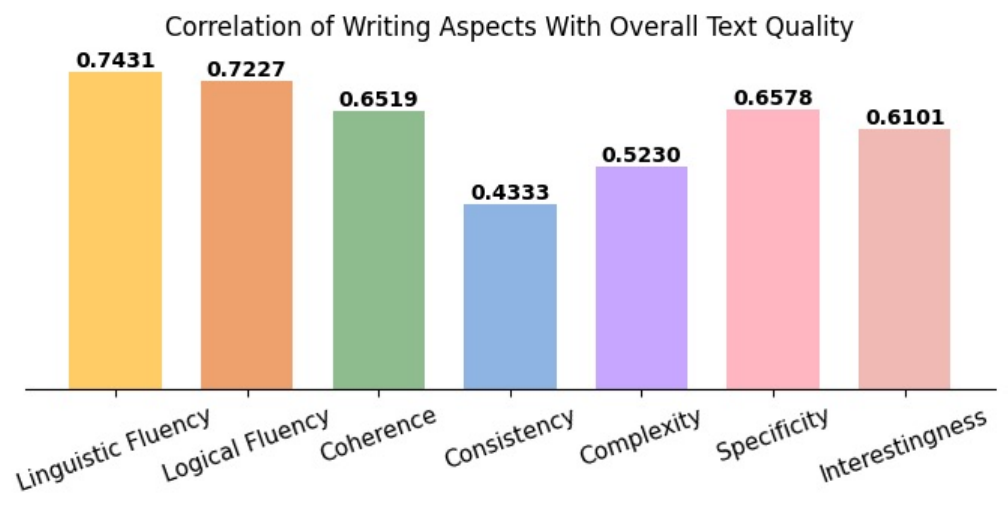}
    \caption{Kendall tau correlations between various writing aspects and overall text quality.}
    \label{fig:writing-quality-corr}
\end{figure}

During the evaluation of the writing quality, writing experts assigned binary overall quality ratings (1 for high quality, 0 for low quality) to the texts. We employed the Kendall tau correlation to examine the relationship between the overall binary ratings and the scores for each evaluation aspect. The analysis (Figure~\ref{fig:writing-quality-corr}) revealed a significant correlation for all the aspects we designed. Interestingly, logical fluency, specificity, and coherence, which were found to be particularly important in determining the perceived quality of written content, are among the aspects that showed the most significant improvement through the WritingPath (Figure~\ref{fig:writing-human-eval-analysis}). 

The progressive improvement in these aspects can be attributed to the effectiveness of using outlines. The initial outline (init) helps organize information more logically and coherently compared to using only metadata (meta), while the augmented outline (aug) further enhances the consistency and richness of the content. These findings highlight the importance of using outlines in the writing process and demonstrate how their gradual enhancement leads to better-structured, more coherent, and content-rich texts, ultimately improving the overall quality of the written output.

\begin{figure*}[ht]
    \centering
    \includegraphics[width=0.7\textwidth]{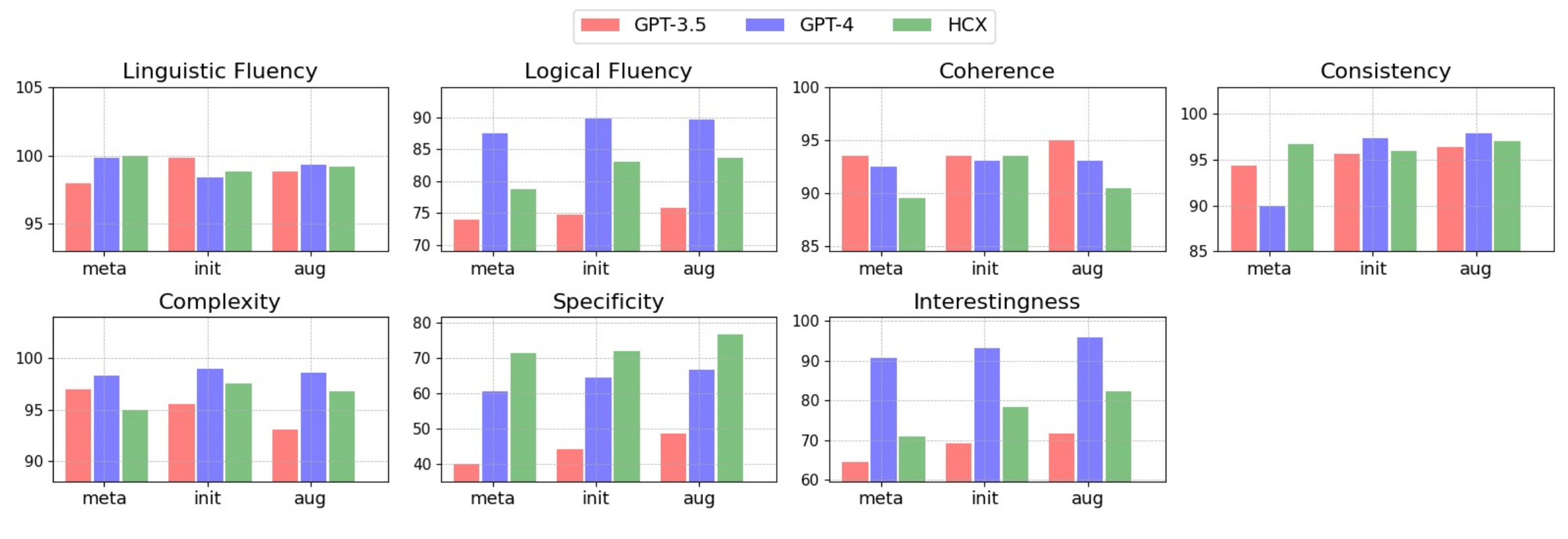}
    \caption{LLM Evaluation}
    \label{fig:writing-eval-analysis-llm}
\end{figure*}

\begin{table*}[h]
    \centering
    \resizebox{0.65\textwidth}{!}{%
       \footnotesize
        \begin{tabular}{l|c|c|cccc}
            \toprule
            \textbf{Model} & \textbf{Category} & \textbf{Outline Type} & \textbf{Logical Alignment} & \textbf{Coherence} & \textbf{Diversity} & \textbf{Repetition} \\
            Metrics & & & NLI ($\uparrow$) & UCI ($\uparrow$) / NPMI ($\uparrow$) & Topic Diversity ($\uparrow$) & Self-BLEU ($\downarrow$) \\
            \midrule
            Eval Level & & & Header-Subheader & Outline & Outline & Outline \\
            \midrule
            \textit{GPT 3.5} & \multirow{2}{*}{Beauty} & Initial & - & 0.638 / 0.298 & 0.488 & 48.01 \\
            & & Augmented & 0.483 & \textbf{1.506} / \textbf{0.553} & \textbf{0.513} & \textbf{23.79} \\ 
            & \multirow{2}{*}{Travel} & Initial & - & 0.835 / 0.454 & \textbf{0.708} & 21.56 \\
            & & Augmented & 0.575 & \textbf{1.646} / \textbf{0.540} & 0.670 & \textbf{13.21} \\
            & \multirow{2}{*}{Gardening} & Initial & - & 0.496 / 0.206 & 0.591 & 26.52 \\
            & & Augmented & 0.658 & \textbf{1.291} / \textbf{0.575} & \textbf{0.592} & \textbf{19.24} \\
            & \multirow{2}{*}{Cooking} & Initial & - & 0.543 / 0.352 & 0.641 & 15.94 \\
            & & Augmented & 0.686 & \textbf{1.003} / \textbf{0.411} & \textbf{0.712} & \textbf{13.71} \\
            & \multirow{2}{*}{IT} & Initial & - & 0.491 / 0.235 & 0.523 & 25.67 \\
            & & Augmented & 0.667 & \textbf{1.180} / \textbf{0.463} & \textbf{0.560} & \textbf{16.69} \\
            \midrule
            \textit{GPT 4} & \multirow{2}{*}{Beauty} & Initial & - & 0.908 / \textbf{0.574} & 0.657 & 36.50 \\
            & & Augmented & 0.577 & \textbf{1.854} / 0.573 & \textbf{0.658} & \textbf{18.80} \\ 
            & \multirow{2}{*}{Travel} & Initial & - & 0.717 / \textbf{0.534} & \textbf{0.691} & 17.61 \\
            & & Augmented & 0.615 & \textbf{1.690} / 0.530 & 0.688 & \textbf{10.63} \\
            & \multirow{2}{*}{Gardening} & Initial & - & 0.833 / 0.398 & 0.676 & 20.33 \\
            & & Augmented & 0.724 & \textbf{1.559} / \textbf{0.555} & \textbf{0.681} & \textbf{13.43} \\
            & \multirow{2}{*}{Cooking} & Initial & - & 0.693 / \textbf{0.468} & 0.720 & 13.85 \\
            & & Augmented & 0.701 & \textbf{1.512} / 0.464 & \textbf{0.745} & \textbf{10.98} \\
            & \multirow{2}{*}{IT} & Initial & - & 0.854 / 0.454 & 0.625 & 16.80 \\
            & & Augmented & 0.702 & \textbf{1.448} / \textbf{0.471} & \textbf{0.633} & \textbf{11.77} \\
            \midrule
            \textit{HyperCLOVA X} & \multirow{2}{*}{Beauty} & Initial & - & 1.030 / \textbf{0.629} & \textbf{0.810} & 22.37 \\
            & & Augmented & 0.504 & \textbf{1.979} / 0.553 & 0.793 & \textbf{12.33} \\
            & \multirow{2}{*}{Travel} & Initial & - & 0.981 / \textbf{0.594} & 0.801 & 11.03 \\
            & & Augmented & 0.626 & \textbf{2.285} / 0.590 & \textbf{0.843} & \textbf{10.20} \\
            & \multirow{2}{*}{Gardening} & Initial & - & 0.694 / 0.280 & 0.623 & 20.73 \\
            & & Augmented & 0.693 & \textbf{1.833} / \textbf{0.563} & \textbf{0.624} & \textbf{12.87} \\
            & \multirow{2}{*}{Cooking} & Initial & - & 0.526 / 0.251 & 0.603 & 17.13 \\
            & & Augmented & 0.774 & \textbf{1.416} / \textbf{0.454} & \textbf{0.658} & \textbf{8.99} \\
            & \multirow{2}{*}{IT} & Initial & - & 0.528 / 0.277 & \textbf{0.606} & 19.22 \\
            & & Augmented & 0.776 & \textbf{1.560} / \textbf{0.536} & 0.596 & \textbf{13.13} \\
            \bottomrule
        \end{tabular}
    }
    \caption{Detailed outline automatic evaluation results.}
    \label{tab:detailed-outline-llm-eval-results}
\end{table*}

\begin{table*}[h]
    \centering
    \resizebox{0.65\textwidth}{!}{%
       \footnotesize
        \begin{tabular}{l|c|c|cccccc}
            \toprule
            \textbf{Model} & \textbf{Category} & \textbf{Outline Type} & \textbf{Cohesion} & \textbf{Natural Flow} & \textbf{Diversity} & \textbf{Redundancy} & \textbf{Usefulness} & \textbf{Improvement} \\
            \midrule
            \textit{GPT 3.5} & \multirow{2}{*}{Beauty} & Initial & \textbf{3.542} & \textbf{2.958} & 2.833 & 2.917 & - & - \\
            & & Augmented & 3.208 & 2.875 & \textbf{3.417} & \textbf{3.375} & 3.000 & 0.542 \\ 
            & \multirow{2}{*}{Travel} & Initial & 3.625 & 2.833 & 3.167 & 2.917 & - & - \\
            & & Augmented & \textbf{3.708} & \textbf{3.125} & \textbf{3.750} & \textbf{3.542} & 3.458 & 0.708 \\
            & \multirow{2}{*}{Gardening} & Initial & 2.958 & 2.375 & \textbf{2.542} & 2.417 & - & - \\
            & & Augmented & \textbf{3.042} & \textbf{2.833} & 3.375 & \textbf{2.667} & 2.542 & 0.708 \\
            & \multirow{2}{*}{Cooking} & Initial & \textbf{3.292} & \textbf{2.417} & 2.417 & 2.583 & - & - \\
            & & Augmented & 2.708 & 2.333 & \textbf{3.458} & \textbf{2.833} & 2.542 & 0.458 \\
            & \multirow{2}{*}{IT} & Initial & \textbf{3.458} & \textbf{2.917} & 2.875 & 2.792 & - & - \\
            & & Augmented & 3.083 & 2.750 & \textbf{3.708} & \textbf{3.250} & 2.875 & 0.542 \\
            \midrule
            \textit{GPT 4} & \multirow{2}{*}{Beauty} & Initial & 3.375 & 2.750 & 3.083 & 3.167 & - & - \\
            & & Augmented & \textbf{3.542} & \textbf{3.208} & \textbf{3.708} & \textbf{3.583} & 3.208 & 0.833 \\ 
            & \multirow{2}{*}{Travel} & Initial & 3.542 & 2.792 & 3.042 & 2.833 & - & - \\
            & & Augmented & \textbf{3.625} & \textbf{3.083} & \textbf{3.792} & \textbf{3.542} & 3.333 & 0.750 \\
            & \multirow{2}{*}{Gardening} & Initial & \textbf{3.792} & 3.125 & 3.083 & 2.917 & - & - \\
            & & Augmented & 3.625 & \textbf{3.208} & \textbf{3.833} & \textbf{3.500} & 3.208 & 0.667 \\
            & \multirow{2}{*}{Cooking} & Initial & \textbf{3.292} & \textbf{2.708} & 2.833 & 2.333 & - & - \\
            & & Augmented & 3.208 & 2.625 & \textbf{3.542} & \textbf{2.917} & 2.833 & 0.542 \\
            & \multirow{2}{*}{IT} & Initial & \textbf{3.000} & \textbf{2.917} & 3.250 & 3.042 & - & - \\
            & & Augmented & \textbf{3.000} & 2.792 & \textbf{3.833} & \textbf{3.583} & 3.042 & 0.750 \\
            \midrule
            \textit{HyperCLOVA X} & \multirow{2}{*}{Beauty} & Initial & 3.375 & 3.292 & 2.583 & 3.125 & - & - \\
            & & Augmented & \textbf{3.500} & \textbf{3.667} & \textbf{3.958} & \textbf{3.833} & 3.667 & 0.917 \\
            & \multirow{2}{*}{Travel} & Initial & \textbf{3.667} & 2.792 & 3.125 & 3.417 & - & - \\
            & & Augmented & 3.583 & \textbf{3.417} & \textbf{4.042} & \textbf{4.000} & 3.542 & 0.833 \\
            & \multirow{2}{*}{Gardening} & Initial & 3.500 & 3.125 & 2.833 & 3.042 & - & - \\
            & & Augmented & \textbf{3.708} & \textbf{3.750} & \textbf{3.958} & \textbf{3.625} & 3.583 & 0.875 \\
            & \multirow{2}{*}{Cooking} & Initial & \textbf{3.500} & 2.958 & 2.750 & 3.250 & - & - \\
            & & Augmented & 3.208 & \textbf{3.375} & \textbf{3.792} & \textbf{3.792} & 3.250 & 0.750 \\
            & \multirow{2}{*}{IT} & Initial & \textbf{3.292} & 2.625 & 2.833 & 3.250 & - & - \\
            & & Augmented & 3.042 & \textbf{3.208} & \textbf{3.917} & \textbf{3.708} & 3.083 & 0.750 \\
            \bottomrule
        \end{tabular}
    }
    \caption{Detailed outline human evaluation results.}
    \label{tab:detailed-outline-human-eval-results}
\end{table*}

\begin{table*}[ht]
    \centering
    \resizebox{0.7\textwidth}{!}{%
       \footnotesize
        \begin{tabular}{l|c|ccccccc|c}
            \toprule
            \textbf{Model} & \textbf{Category} & \textbf{Linguistic Fl.} & \textbf{Logical Fl.} & \textbf{Coh.} & \textbf{Cons.} & \textbf{Comple.} & \textbf{Spec.} & \textbf{Int.} & \textbf{Overall} \\
            \midrule
            \textit{GPT 3.5} & Beauty & 96.00 & 75.42 & 90.00 & 94.17 & 89.58 & 52.50 & 63.06 & 80.10 \\
            & Travel & 100.00 & 78.54 & 97.08 & 96.94 & 97.08 & 65.83 & 81.67 & 88.16 \\
            & Gardening & 98.67 & 75.63 & 95.00 & 96.11 & 89.17 & 49.17 & 78.89 & 83.23 \\
            & Cooking & 99.50 & 72.92 & 95.42 & 97.78 & 95.00 & 35.83 & 84.17 & 82.94 \\
            & IT & 100.00 & 76.67 & 97.50 & 96.94 & 94.17 & 39.17 & 50.83 & 79.33 \\
            \midrule
            & Total & 98.83 & 75.83 & 95.00 & 96.39 & 93.00 & 48.50 & 71.72 & 82.75 \\
            \midrule
            \textit{GPT 4} & Beauty & 99.17 & 89.79 & 97.50 & 99.44 & 99.17 & 84.58 & 98.61 & 95.47 \\
            & Travel & 99.00 & 90.21 & 91.67 & 97.22 & 96.67 & 70.00 & 96.94 & 91.67 \\
            & Gardening & 99.67 & 90.00 & 94.17 & 98.33 & 100.00 & 74.17 & 97.78 & 93.44 \\
            & Cooking & 99.67 & 89.58 & 93.75 & 98.61 & 97.92 & 63.33 & 96.67 & 91.36 \\
            & IT & 99.17 & 88.96 & 88.33 & 96.11 & 99.17 & 41.67 & 76.11 & 84.22 \\
            \midrule
            & Total & \textbf{99.33} & \textbf{89.71} & \textbf{93.08} & \textbf{97.94} & \textbf{98.58} & 66.75 & \textbf{93.22} & \textbf{91.23} \\
            \midrule
            \textit{HyperCLOVA X} & Beauty & 100.00 & 88.33 & 98.33 & 100.00 & 90.00 & 90.42 & 91.38 & 94.07 \\
            & Travel & 99.50 & 83.75 & 90.42 & 97.50 & 91.25 & 79.58 & 92.77 & 90.68 \\
            & Gardening & 99.33 & 88.13 & 93.33 & 98.61 & 95.00 & 70.42 & 84.16 & 89.85 \\
            & Cooking & 98.67 & 82.29 & 88.75 & 96.67 & 90.42 & 87.92 & 91.11 & 90.83 \\
            & IT & 98.50 & 76.04 & 81.25 & 92.50 & 87.08 & 55.00 & 52.50 & 77.55 \\
            \midrule
            & Total & 99.20 & 83.71 & 90.42 & 97.06 & 90.75 & \textbf{76.67} & 82.38 & 88.60 \\
            \bottomrule
        \end{tabular}
    }
    \caption{Detailed writing LLM evaluation results.}
    \label{tab:detailed-writing-llm-eval-results}
\end{table*}

\begin{table*}[h]
    \centering
    \resizebox{0.7\textwidth}{!}{%
       \footnotesize
        \begin{tabular}{l|c|ccccccc|c}
            \toprule
            \textbf{Model} & \textbf{Category} & \textbf{Linguistic Fl.} & \textbf{Logical Fl.} & \textbf{Coh.} & \textbf{Cons.} & \textbf{Comple.} & \textbf{Spec.} & \textbf{Int.} & \textbf{Overall} \\
            \midrule
            \textit{GPT 3.5} & Beauty & 51.43 & 30.29 & 51.43 & 92.57 & 67.14 & 14.29 & 29.29 & 48.06 \\
            & Travel & 68.00 & 56.00 & 68.57 & 87.43 & 82.14 & 47.14 & 55.71 & 66.43 \\
            & Gardening & 45.14 & 30.86 & 44.29 & 87.43 & 52.86 & 18.57 & 30.00 & 44.16 \\
            & Cooking & 46.29 & 8.86 & 21.43 & 80.00 & 72.86 & 7.14 & 32.86 & 38.49 \\
            & IT & 47.43 & 29.71 & 45.71 & 93.14 & 57.14 & 18.57 & 27.86 & 45.65 \\
            \midrule
            & Total & 51.66 & 31.14 & 46.29 & 88.11 & 66.43 & 21.14 & 35.14 & 48.56 \\
            \midrule
            \textit{GPT 4} & Beauty & 72.00 & 67.43 & 82.86 & 93.71 & 86.43 & 56.43 & 72.14 & 75.86 \\
            & Travel & 76.00 & 71.43 & 82.86 & 89.71 & 86.43 & 67.14 & 79.29 & 78.98 \\
            & Gardening & 66.29 & 61.14 & 75.71 & 85.71 & 70.71 & 54.29 & 61.43 & 67.90 \\
            & Cooking & 63.43 & 50.86 & 61.43 & 90.86 & 82.14 & 47.14 & 62.14 & 65.43 \\
            & IT & 62.29 & 52.00 & 61.43 & 86.29 & 75.71 & 45.71 & 56.43 & 62.84 \\
            \midrule
            & Total & 68.00 & 60.57 & 72.86 & 89.26 & 80.29 & 54.14 & 66.29 & 70.20 \\
            \midrule
            \textit{HyperCLOVA X} & Beauty & 92.00 & 87.43 & 91.43 & 99.43 & 92.14 & 69.29 & 81.43 & 87.59 \\
            & Travel & 95.43 & 91.43 & 95.71 & 99.43 & 100.00 & 84.29 & 85.00 & 93.04 \\
            & Gardening & 88.57 & 85.71 & 97.14 & 99.43 & 95.71 & 75.71 & 82.86 & 89.31 \\
            & Cooking & 90.86 & 85.71 & 90.00 & 98.29 & 96.43 & 81.43 & 82.14 & 89.27 \\
            & IT & 81.71 & 72.00 & 81.43 & 93.71 & 78.57 & 59.29 & 68.57 & 76.47 \\
            \midrule
            & Total & \textbf{89.71} & \textbf{84.46} & \textbf{91.14} & \textbf{98.06} & \textbf{92.57} & \textbf{74.00} & \textbf{80.00} & \textbf{87.13} \\
            \bottomrule
        \end{tabular}
    }
    \caption{Detailed writing human evaluation results.}
    \label{tab:detailed-writing-human-eval-results}
\end{table*}

\definecolor{LingFluColor}{HTML}{ffcc66} 
\definecolor{LogicFluColor}{HTML}{eea16c} 
\definecolor{CohColor}{HTML}{8fbc8f} 
\definecolor{ConsColor}{HTML}{8eb4e3} 
\definecolor{CompleColor}{HTML}{c6a6ff} 
\definecolor{SpecColor}{HTML}{ffb6c1} 
\definecolor{IntColor}{HTML}{F1B9B3} 

\begin{table*}[h]
    \centering
    \footnotesize
    \arrayrulecolor{black}
    \begin{tabular}{c|c|l}
        \toprule
        \textbf{Aspect} & \textbf{Subaspect} & \textbf{Descriptions} \\ 

        \hhline{>{\arrayrulecolor{black}}===}
        {\cellcolor{LingFluColor}} & {\cellcolor{LingFluColor}}Natural Expression & \begin{tabular}[c]{@{}p{0.6\linewidth}p{0.2\textwidth}@{}}\tiny Does the given text read naturally without any unnatural rhythm or excessively emphasized parts?\\\tiny 주어진 글이 부자연스러운 리듬이나 과도하게 강조된 부분 없이 자연스럽게 읽히나요?\end{tabular} \\ 
        \hhline{|>{\arrayrulecolor{LingFluColor}}->{\arrayrulecolor{black}}--|}
        {\cellcolor{LingFluColor}} & {\cellcolor{LingFluColor}}Text Length & \begin{tabular}[c]{@{}p{0.6\linewidth}@{}}\tiny Is the length of the text suitable for the purpose and is it not excessively verbose or overly concise?\\\tiny 텍스트의 길이가 목적에 적합하며 과도하게 장황하거나 지나치게 간결하지는 않은 글인가요?\end{tabular} \\ 
        \hhline{>{\arrayrulecolor{LingFluColor}}->{\arrayrulecolor{black}}--|}
        {\cellcolor{LingFluColor}} & {\cellcolor{LingFluColor}}Vocabulary & \begin{tabular}[c]{@{}p{0.6\linewidth}@{}}\tiny Is the vocabulary appropriate for the context, not overly complex, and suitable for the topic and reader?\\\tiny 어휘가 맥락에 맞지 않거나 지나치게 복잡하지 않고, 주제와 독자에 적합한가요?\end{tabular} \\ 
        \hhline{>{\arrayrulecolor{LingFluColor}}->{\arrayrulecolor{black}}--|}
        {\cellcolor{LingFluColor}} & {\cellcolor{LingFluColor}}Syntax & \begin{tabular}[c]{@{}p{0.6\linewidth}@{}}\tiny Is the composition and sentence structure of the given text correct?\\\tiny 주어진 글의 구성과 문장의 구조가 올바른가요?\end{tabular} \\ 
        \hhline{>{\arrayrulecolor{LingFluColor}}->{\arrayrulecolor{black}}--|}
        \multirow{-9}{*}{{\cellcolor{LingFluColor}}Linguistic Fluency} & {\cellcolor{LingFluColor}}Mechanic-Spelling, Punctuation & \begin{tabular}[c]{@{}p{0.6\linewidth}@{}}\tiny Is the spelling and punctuation of the given text correctly applied?\\\tiny 주어진 글의 철자와 문장부호가 올바르게 적용되었나요?\end{tabular} \\ 

        \hhline{>{\arrayrulecolor{black}}---|}
        \cellcolor{LogicFluColor} & {\cellcolor{LogicFluColor}}Organization (layout) & \begin{tabular}[c]{@{}p{0.6\linewidth}@{}}\tiny Does the given text have a clear and effective structure (layout)?\\\tiny 주어진 글은 명확하고 효과적인 구조 (레이아웃)를 가지고 있나요?\end{tabular} \\ 
        \hhline{>{\arrayrulecolor{LogicFluColor}}->{\arrayrulecolor{black}}--|}
        \cellcolor{LogicFluColor} & {\cellcolor{LogicFluColor}}Repetitive Content & \begin{tabular}[c]{@{}p{0.6\linewidth}@{}}\tiny Is the text free of repetitive or unnecessary content?\\\tiny 텍스트 내에서 반복되는 내용이나 불필요한 내용이 없는 글인가요?\end{tabular} \\ 
        \hhline{>{\arrayrulecolor{LogicFluColor}}->{\arrayrulecolor{black}}--|}
        \cellcolor{LogicFluColor} & {\cellcolor{LogicFluColor}} & \begin{tabular}[c]{@{}p{0.6\linewidth}@{}}\tiny In the text, are the sentences well connected and progressing naturally and logically?\\\tiny 글 내에서 문장들이 잘 연결되어 있어 자연스럽고 논리적으로 진행이 되나요 ?\end{tabular} \\ 
        \hhline{>{\arrayrulecolor{LogicFluColor}}-->{\arrayrulecolor{black}}|-|}
        \cellcolor{LogicFluColor} & \multirow{-3}{*}{{\cellcolor{LogicFluColor}}Inter-sentence Cohesion} & \begin{tabular}[c]{@{}p{0.6\linewidth}@{}}\tiny Did you use conjunctions appropriately to improve readability?\\\tiny 가독성을 높이기 위한 접속사를 적절하게 사용했나요?\end{tabular} \\ 
        \hhline{>{\arrayrulecolor[rgb]{0.8,0.8,0.8}}|>{\arrayrulecolor{LogicFluColor}}->{\arrayrulecolor{black}}--|}
        \multirow{-9}{*}{{\cellcolor{LogicFluColor}}Logical Fluency} & {\cellcolor{LogicFluColor}}Inter-paragraph Cohesion & \begin{tabular}[c]{@{}p{0.6\linewidth}@{}}\tiny Are the paragraphs in the text logically connected and progressing with each other?\\\tiny 텍스트 내의 단락들이 논리적으로 연결되어 서로 진행되나요?\end{tabular} \\ 

        \hhline{>{\arrayrulecolor{black}}---|}
        {\cellcolor{CohColor}} & {\cellcolor{CohColor}}Topic Consistency & \begin{tabular}[c]{@{}p{0.6\linewidth}@{}}\tiny Is the entire article consistently progressing with the central theme as the focus?\\\tiny 전체 글이 중심 주제를 중심으로 일관되게 진행되나요?\end{tabular} \\ 
        \hhline{>{\arrayrulecolor{CohColor}}->{\arrayrulecolor{black}}--|}
        \multirow{-3}{*}{{\cellcolor{CohColor}}Coherence} & {\cellcolor{CohColor}}Topic Setence and Paragraph & \begin{tabular}[c]{@{}p{0.6\linewidth}@{}}\tiny Does each paragraph of the article have a clear subtopic centered around the main idea?\\\tiny 글의 각 문단이 주요 아이디어를 중심으로 명확한 소주제를 가지고 있나요?\end{tabular} \\ 

        \hhline{>{\arrayrulecolor{black}}---|}
        {\cellcolor{ConsColor}} & {\cellcolor{ConsColor}} & \begin{tabular}[c]{@{}p{0.6\linewidth}@{}}\tiny Is a consistent narrative tone and style maintained throughout the entire text?\\\tiny 텍스트 전체에서 일관된 서술 어조와 어투가 유지되나요?\end{tabular} \\ 
        \hhline{>{\arrayrulecolor{ConsColor}}-->{\arrayrulecolor{black}}-|}
        {\cellcolor{ConsColor}} & \multirow{-3}{*}{{\cellcolor{ConsColor}}Tone} & \begin{tabular}[c]{@{}p{0.6\linewidth}@{}}\tiny Is there no sudden change in tone in the context of the writing?\\\tiny 글의 맥락에서 급격한 어조 변화가 없는 글인가요?\end{tabular} \\ 
        \hhline{>{\arrayrulecolor{ConsColor}}->{\arrayrulecolor{black}}--|}
        {\cellcolor{ConsColor}} & {\cellcolor{ConsColor}}Stance/Posture & \begin{tabular}[c]{@{}p{0.6\linewidth}@{}}\tiny Does the author present a consistent opinion on the topic in the writing? (Should not present conflicting opinions on the same subject)\\\tiny 저자는 글에서 주제에 대한 일관된 의견을 제시하나요? (동일한 대상에 대한 상반된 의견을 제시하지 않아야 함)\end{tabular} \\ 
        \hhline{>{\arrayrulecolor{ConsColor}}->{\arrayrulecolor{black}}--|}
        {\cellcolor{ConsColor}} & {\cellcolor{ConsColor}} & \begin{tabular}[c]{@{}p{0.6\linewidth}@{}}\tiny Does the given text maintain a consistent style type (spoken language, written language, informal, formal, etc.)?\\\tiny 주어진 글이 일관된 스타일의 유형 (구어체, 문어체, 반말, 존댓말 등의 유형)을 유지하나요?\end{tabular} \\ 
        \hhline{>{\arrayrulecolor{ConsColor}}-->{\arrayrulecolor{black}}-|}
        \multirow{-9}{*}{{\cellcolor{ConsColor}}Consistency} & \multirow{-3}{*}{{\cellcolor{ConsColor}}Style} & \begin{tabular}[c]{@{}p{0.6\linewidth}@{}}\tiny Do you consistently use abbreviations and acronyms when necessary?\\\tiny 필요 시 약어와 머리글자가 일관되게 사용되나요?\end{tabular} \\ 

        \hhline{>{\arrayrulecolor{black}}---|}
        {\cellcolor{CompleColor}} & {\cellcolor{CompleColor}} & \begin{tabular}[c]{@{}p{0.6\linewidth}@{}}\tiny Is it a clear text that does not excessively use uncommon or complex words?\\\tiny 일반적이지 않거나 복잡한 단어들이 과도하게 등장하지 않는 명료한 글인가요?\end{tabular} \\ 
        \hhline{>{\arrayrulecolor{CompleColor}}-->{\arrayrulecolor{black}}-|}        
        {\cellcolor{CompleColor}} & \multirow{-3}{*}{{\cellcolor{CompleColor}}Vocabulary} & \begin{tabular}[c]{@{}p{0.6\linewidth}@{}}\tiny Is the definition of unfamiliar and difficult words provided and are they used appropriately in context?\\\tiny 낯설고 어려운 단어에 대한 정의가 되어 있고 문맥에 맞게 잘 사용되었나요?\end{tabular} \\ 
        \hhline{>{\arrayrulecolor{CompleColor}}->{\arrayrulecolor{black}}--|}
        {\cellcolor{CompleColor}} & {\cellcolor{CompleColor}} & \begin{tabular}[c]{@{}p{0.6\linewidth}@{}}\tiny Is the given text clearly structured without excessively complex sentence structures?\\\tiny 주어진 글이 과도하게 복잡한 문장 구조를 가진 문장들 없이 명확하게 구성되어 있나요?\end{tabular} \\ 
        \hhline{>{\arrayrulecolor{CompleColor}}-->{\arrayrulecolor{black}}-|}
        \multirow{-7}{*}{{\cellcolor{CompleColor}}Complexity} & \multirow{-3}{*}{{\cellcolor{CompleColor}}Syntax} & \begin{tabular}[c]{@{}p{0.6\linewidth}@{}}\tiny Do the first sentences of each paragraph start differently? (Asking if the text has paragraphs that do not all start the same way)\\\tiny 각 문단의 첫 문장이 다양하게 시작되나요? (각 문단의 시작이 모두 동일하지 않은 글인지 질문)\end{tabular} \\ 

        \hhline{>{\arrayrulecolor{black}}---|}
        \cellcolor{SpecColor} & {\cellcolor{SpecColor}} & \begin{tabular}[c]{@{}p{0.6\linewidth}@{}}\tiny Is the example appropriately connected to the topic of the article?\\\tiny 예시가 글의 주제와 적절하게 연결되어 있나요?\end{tabular} \\ 
        \hhline{>{\arrayrulecolor{SpecColor}}-->{\arrayrulecolor{black}}-|}
        \cellcolor{SpecColor} & \multirow{-3}{*}{{\cellcolor{SpecColor}}Use of Examples and Review} & \begin{tabular}[c]{@{}p{0.6\linewidth}@{}}\tiny Was the author's personal experience mentioned specifically?\\\tiny 작성자의 개인적인 경험이 구체적으로 언급되었나요?\end{tabular} \\ 
        \hhline{>{\arrayrulecolor{SpecColor}}->{\arrayrulecolor{black}}--|}
        \cellcolor{SpecColor} & {\cellcolor{SpecColor}} & \begin{tabular}[c]{@{}p{0.6\linewidth}@{}}\tiny In the writing, were specific numerical values such as ratios and quantities mentioned?\\\tiny 글에서 구체적으로 비율, 수량과 같은 수치들이 언급되었나요?\end{tabular} \\ 
        \hhline{>{\arrayrulecolor{SpecColor}}-->{\arrayrulecolor{black}}-|}
        \multirow{-7}{*}{{\cellcolor{SpecColor}}Specificity} & \multirow{-3}{*}{{\cellcolor{SpecColor}}Detailed Descriptions} & \begin{tabular}[c]{@{}p{0.6\linewidth}@{}}\tiny When introducing details in a writing, do you appropriately utilize context or background information?\\\tiny 글에서 세부 사항을 소개할 때 맥락이나 배경 정보를 적절하게 활용하나요?\end{tabular} \\ 

        \hhline{>{\arrayrulecolor{black}}---|}
        \cellcolor{IntColor} & {\cellcolor{IntColor}}Engagement & \begin{tabular}[c]{@{}p{0.6\linewidth}@{}}\tiny Was the blog post written based on an appealing storytelling approach? (It's okay if an exaggerated tone is included)\\\tiny 블로그 글이 매력적인 스토리텔링 접근 방식을 기반으로 작성되었나요? (과장된 어조가 포함되어도 괜찮음)\end{tabular} \\ 
        \hhline{>{\arrayrulecolor{IntColor}}->{\arrayrulecolor{black}}--|}
        \cellcolor{IntColor} & {\cellcolor{IntColor}}Kindness & \begin{tabular}[c]{@{}p{0.6\linewidth}@{}}\tiny Was the written blog post written in a friendly tone for the readers?\\\tiny 작성된 블로그 글은 독자들에게 친근한 어조로 작성되었나요?\end{tabular} \\ 
        \hhline{>{\arrayrulecolor{IntColor}}->{\arrayrulecolor{black}}--|}
        \cellcolor{IntColor} & {\cellcolor{IntColor}} & \begin{tabular}[c]{@{}p{0.6\linewidth}@{}}\tiny Does the written blog post include the author's unique ideas or perspectives?\\\tiny 작성된 블로그 글에는 작성자의 독특한 아이디어나 관점이 포함되어 있나요?\end{tabular} \\ 
        \hhline{>{\arrayrulecolor{IntColor}}->{\arrayrulecolor{IntColor}}->{\arrayrulecolor{black}}-|}
        \multirow{-7}{*}{{\cellcolor{IntColor}}Interestingness} & \multirow{-3}{*}{{\cellcolor{IntColor}}Originality} & \begin{tabular}[c]{@{}p{0.6\linewidth}@{}}\tiny Does the writer's personal experience add freshness to the writing?\\\tiny 작성자의 개인적인 경험이 글에 신선함을 더하나요?\end{tabular} \\
        \hhline{>{\arrayrulecolor{black}}---}
        \bottomrule
    \end{tabular}
    \arrayrulecolor{black}
    \caption{Evaluation principles.}
    \label{tab:eval-principles}
\end{table*}

\begin{figure*}[h]
    \centering
    \begin{tcolorbox}[
    colback=yellow!10!white,      
    colframe=black,               
    colbacktitle=yellow!60!white, 
    coltitle=black,               
    rounded corners,              
    boxrule=0.8mm,                
    width=\textwidth,             
    fonttitle=\large,
    title={\textbf{Writing Evaluation Prompt}}, 
]
        \texttt{You will be given one text written for a blog post.\\
        Your task is to rate the written text on one metric.\\
        Please read and understand these instructions carefully.\\
        Keep this document open while reviewing and refer to it as needed.
        You are a writing expert! it is crucial to apply a robust evaluation.\\
        \\ \#\# Evaluation Criteria:\\\{aspect\} - \{definition\}\\
        \#\#\# Guidelines\#\#\#\\
        1. Read these guidelines completely.\\
        2. Read the Written Text attentively.\\
        3. Comprehend the questions and the meaning of the \{aspect\}.\\
        4. Answer each question with 'yes' or 'no', without any explanations.\\
        5. Use the prescribed answer format.\\
        \\
        \#\#\# Output Format\#\#\#\\
        Q: [Question] A: [Answer]\\
        Q: [Question] A: [Answer]\\
        ...\\
        \\
        \#\#\# Questions\#\#\#\\
        Q. \{question\}\\
        \\
        Blog text: \{writing\}\\
        \\
        Your Answers:\\}
\end{tcolorbox}
    \caption{Writing Evaluation Prompt for Checklist-based Assessment.}
    \label{fig:writing-evaluation-prompt}
\end{figure*}

\begin{figure*}[h]
    \centering
    \begin{tcolorbox}[
    colback=green!10!white,      
    colframe=black,               
    colbacktitle=green!50!white, 
    coltitle=black,               
    rounded corners,              
    boxrule=0.8mm,                
    width=\textwidth,             
    fonttitle=\large,
    title={\textbf{WritingPath Prompt}}, 
]
        \textbf{Prompt for Metadata construction (step \#1): }\\
        \texttt{We aim to systematically organize blog posts by dividing them into four categories: \\
        1. the purpose of the post\\ 2. the type of post\\ 3. the style of the post\\ 4. keywords. \\
        An example of the expected format is provided below.\\
        \\
        \{examples\}\\
        \\
        Similar to the example provided, please categorize the blog post below in detail according to \\
        1. purpose, 2. type, 3. style, and 4. keywords, where keywords are composed of words.\\
        \\
        ==Blog post==\\
        \{original blog text\}\\ }
        
        \textbf{Prompt for Generation of Title and Initial Outline (step \#2): }\\
        \texttt{Based on the metadata, I plan to create the title and a simple table of contents for the article. \\
        Below is an example of the desired format.\\
        \\
        \{example\}\\
        \\
        Following the example above, based on the post information provided below,
        only create "==Title==" and a brief "==Initial Outline==". \\
        Do not generate an excessively long table of contents. \\
        The table of contents should not be a simple list; \\
        do not write it in paragraph form. Do not create subheadings. \\
        Only the title and table of contents should be generated. \\
        The table of contents must be numbered in sequence. \\
        You must strictly follow the format for the title and table of contents below.\\
        \\
        ==Meta data==\\
        \{meta data\} \\}
\end{tcolorbox}
    \caption{WritingPath Prompt for Each Stage (Step 1 and 2).}
    \label{fig:writing-path-prompt-1}
\end{figure*}

\begin{figure*}[h]
    \centering
    \begin{tcolorbox}[
    colback=green!10!white,      
    colframe=black,               
    colbacktitle=green!50!white, 
    coltitle=black,               
    rounded corners,              
    boxrule=0.8mm,                
    width=\textwidth,             
    fonttitle=\large,
    title={\textbf{WritingPath Prompt}}, 
]
    \textbf{Prompt for Generation of Augmented Outline (step \#4): } \\ 
    \texttt{Map the necessary additional information below to create an augmented outline. Here is an example.\\
        \\
        \{example\}\\
        \\
        Following the method above, create an ==Augmented Outline==. \\
        Specifically, incorporate new information as subheadings under the existing headings, \\ ensuring that each heading and its subheadings are themed consistently.\\
        \\
        ==Additional Information==\\
        \{additional information from browsing\}\\
        ==Initial Outline==\\
        \{initial outline\} \\}
        
        \textbf{Prompt for Generation of Text (step \#5): }\\ 
        \texttt{Based on the title and current table of contents below, \\
        I plan to write the {i + 1}th paragraph suitable for a blog post. \\
        Writing should naturally follow the flow of the post information and the augmented outline.\\
        Write in a friendly and attractive tone like bloggers, making it interesting for the reader. \\
        The written content should be engaging and captivating for the reader.\\
        \\
        ==Augmented Outline==\\
        \{augmented outline\}\\
        ==Meta Data==\\
        \{meta data\}\\  
        \\
        Below are the title and current table of contents for writing the blog post.\\
        ==Title==\\
        \{title\}\\
        ==Current Outline==\\
        \{current section\} \\}
\end{tcolorbox}
    \caption{WritingPath Prompt for Each Stage (Step 4 and 5).}
    \label{fig:writing-path-prompt-2}
\end{figure*}

\end{document}